\pdfoutput=1

\documentclass[11pt]{article}

\usepackage[final]{coling}

\usepackage{times}
\usepackage{latexsym}

\usepackage[T1]{fontenc}

\usepackage[utf8]{inputenc}

\usepackage{microtype}

\usepackage{inconsolata}

\usepackage{graphicx}

\usepackage{amsmath}
\usepackage{booktabs}

\usepackage{amsthm}
\usepackage{amsmath}
\usepackage{amssymb}

\usepackage{helvet}  
\usepackage{courier}  
\usepackage{graphicx} 
\usepackage{caption} 
\usepackage{subfigure}
\usepackage{amsfonts}
\usepackage{multirow}
\usepackage{color}

\usepackage{algorithm}
\usepackage{algpseudocode}
\usepackage{arydshln}

\usepackage{color}
\usepackage{xcolor}

\newcommand{\gray}[1]{{\color{gray}#1}}

\algnewcommand{\algorithmicforeach}{\textbf{for each}}
\algdef{SE}[FOR]{ForEach}{EndForEach}[1]
{\algorithmicforeach\ #1\ \algorithmicdo}
{\algorithmicend\ \algorithmicforeach}

\title{FedMKT: Federated Mutual Knowledge Transfer for Large and Small Language Models}

\author{
 \textbf{Tao Fan\textsuperscript{1, 2}},
 \textbf{Guoqiang Ma\textsuperscript{2}},
 \textbf{Yan Kang\textsuperscript{2}},
 \textbf{Hanlin Gu\textsuperscript{2}},
 \textbf{Yuanfeng Song\textsuperscript{2}},
 \\
 \textbf{Lixin Fan\textsuperscript{2}},
 \textbf{Kai Chen\textsuperscript{1}},
 \textbf{Qiang Yang\textsuperscript{1,2}}
\\
 \textsuperscript{1} The Hong Kong University of Science and Technology, Hong Kong, China
 \\
 \textsuperscript{2} WeBank Co., Ltd, Shenzhen, China
\\
 \small{
   \textbf{Correspondence:} \href{mailto:tfanac@cse.ust.hk}{tfanac@cse.ust.hk}
 }
}

\begin{document}
\maketitle

\begin{abstract}
Recent research in federated large language models (LLMs) has primarily focused on enabling clients to fine-tune their locally deployed homogeneous LLMs collaboratively or on transferring knowledge from server-based LLMs to small language models (SLMs) at downstream clients. However, a significant gap remains in the simultaneous mutual enhancement of both the server's LLM and clients' SLMs. To bridge this gap, we propose FedMKT, a parameter-efficient federated mutual knowledge transfer framework for large and small language models. This framework is designed to adaptively transfer knowledge from the server's LLM to clients' SLMs while concurrently enhancing the LLM with clients' unique domain insights. We facilitate token alignment using minimum edit distance (MinED) and then selective mutual knowledge transfer between client-side SLMs and a server-side LLM, aiming to collectively enhance their performance.
Through extensive experiments across three distinct scenarios, we evaluate the effectiveness of FedMKT by utilizing diverse public LLMs and SLMs on a variety of NLP text generation tasks. Empirical results demonstrate that FedMKT simultaneously boosts the performance of both LLMs and SLMs.
Our code has been contributed to the FATE open-source project and is now publicly accessible at \textit{\url{https://github.com/FederatedAI/FATE-LLM/tree/main/python/fate_llm/algo/fedmkt}}

\end{abstract}

\section{Introduction} 
The emergence of Large Language Models (LLMs) has marked a revolutionary shift in artificial intelligence, significantly transforming our understanding of natural language processing capabilities. The advent of cutting-edge LLMs like ChatGPT ~\cite{Chatgpt}, Gemma2~\cite{team2024gemma}, and LLaMa2~\cite{touvron2023llama} with their billions of parameters, has sparked the imagination of both researchers and practitioners, owing to their exceptional performance across diverse text generation tasks. 
Despite their widespread success in various general NLP tasks, LLMs face challenges that hinder their adoption in domain-specific applications~\cite{kang2023grounding, fan2023fate, fan2024pdss}. The primary challenges include domain-specific knowledge Privacy, constrained computing resources, and mutual knowledge transfer between the LLM and SLMs. A significant challenge arises from the inherent model heterogeneity between the LLM and SLMs, particularly when aligning distributions of output logits. The mismatch between the tokenizers of different LLM and SLMs poses a notable obstacle. Furthermore, the mutual transfer of knowledge between the server's LLM and clients' SLMs remains a largely unexplored area in academic literature, warranting further investigation.

To fill these gaps, we propose FedMKT, a novel federated mutual knowledge transfer framework designed to improve the performance of both large and small language models. By leveraging the complementary strengths of federated learning and knowledge distillation, FedMKT facilitates effective mutual knowledge transfer between clients' SLMs and the LLM owned by the server.

\begin{figure*}[htb]
  \centering
  \includegraphics[width=0.98\linewidth]{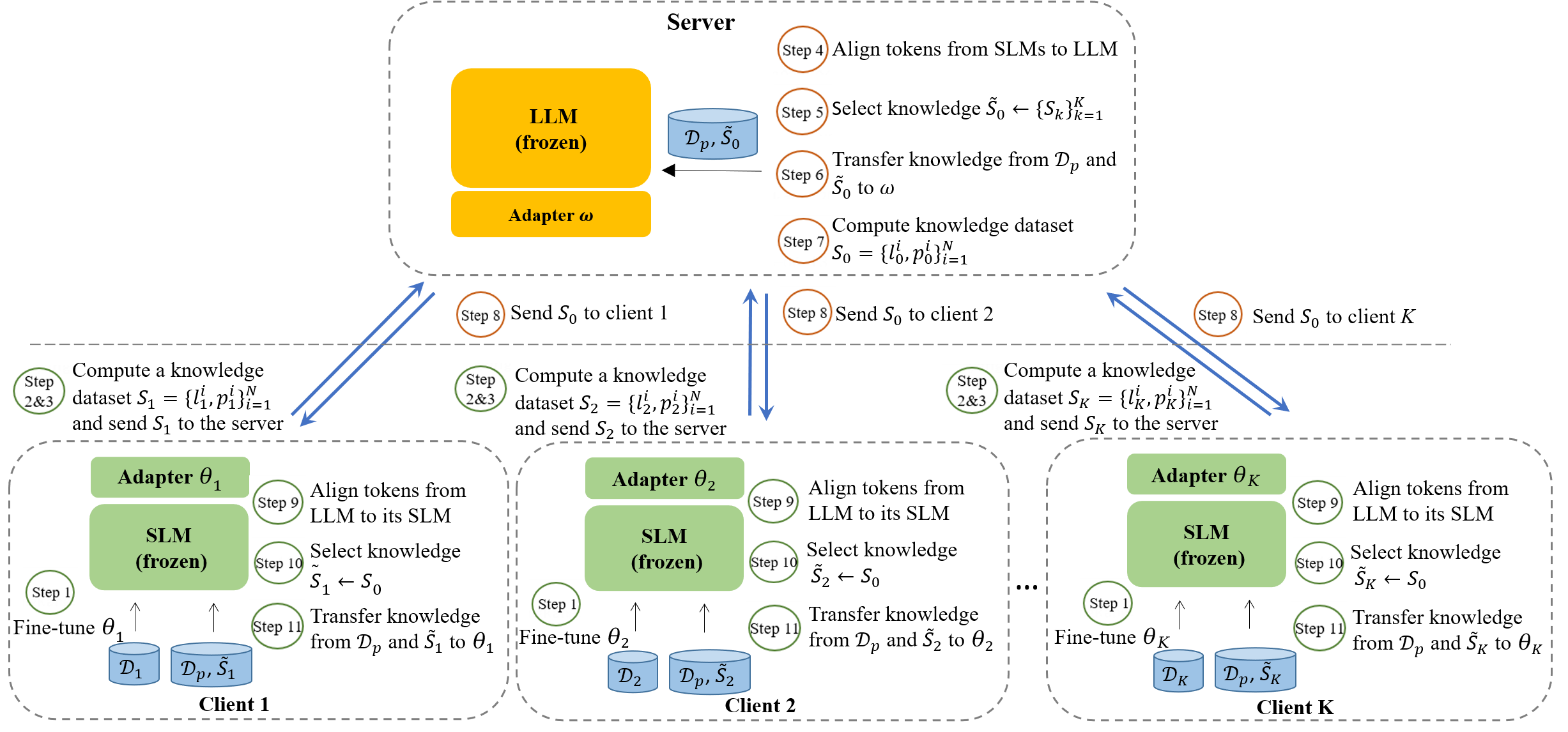}
  \caption{
  Overview of the proposed FedMKT workflow. Each communication round of FedMKT involves 11 steps to fine-tune the server' LLM and clients' SLMs.
  }
  \label{fig:fedmkt}
\end{figure*}

As illustrated in Figure \ref{fig:fedmkt}, FedMKT deploys an LLM on the server and a set of $K$ heterogeneous SLMs across various clients. The cornerstone of FedMKT lies in its selective mutual knowledge transfer process. During each round of federated learning, the clients transmit the output logits of their updated SLMs on the public dataset to the server. Subsequently, the server selectively aggregates and extracts the knowledge encoded within these SLMs output logits into the server-side LLM. This process allows the server LLM to incorporate the domain-specific knowledge learned by the clients, thereby enhancing its comprehensive capabilities. Simultaneously, the server-side LLM also selectively distills its knowledge to the clients’ SLMs, which is similar to the knowledge transfer from clients to the server. By leveraging the knowledge of the server LLM, the clients’ SLMs are able to improve their performance and generalize better to unseen data. To address the model heterogeneity between the LLM and SLMs, FedMKT incorporates a token alignment technique utilizing minimum edit distance (MinED) prior to knowledge transfer. This alignment ensures seamless integration and efficient knowledge transfer between LLM and SLMs.

Our contributions are summarized as follows:
\begin{itemize}

\item \textbf{Federated Mutual Knowledge Transfer Framework}. FedMKT introduces a novel federated mutual knowledge transfer framework that enables effective knowledge transfer between an LLM deployed on the server and SLMs residing on clients. This framework fills the gap by simultaneously enhancing both the server's LLM and the clients' SLMs. Our work is tailored for text generation tasks within the context of LLMs and supports both Heterogeneous and Homogeneous scenarios between client SLMs. To our best knowledge, our work is the first published research in this field.

\item \textbf{Selective Knowledge Transfer and Token Alignment}.  FedMKT implements a selective knowledge transfer mechanism that selectively distills knowledge from the most informative SLMs to the server's LLM and vice versa. Furthermore, it incorporates a token alignment technique using minimum edit distance (MinED) to address model heterogeneity between LLM and SLMs, ensuring efficient knowledge transfer.

\item \textbf{Empirical Evaluation and Performance Enhancement}. 
Extensive experiments conducted based on various publicly available LLMs and SLMs, have shown that FedMKT exhibits competitive performance across a broad spectrum of NLP text-generation tasks.
We evaluate FedMKT with Heterogeneous, Homogeneous, and One-to-One settings. The results show that the performance of SLMs can be significantly enhanced with the help of the LLM, while the LLM can deliver comparable results to fine-tuning with all clients' data centralized. 

\end{itemize}

\section{Related Work}
\subsection{Model Heterogeneous Federated Learning}
 
Model heterogeneous federated learning (MHFL) aims to address the challenges associated with heterogeneity in federated learning (FL)~\cite{yang2019federated, mcmahan2017communication, liu2021fate, cheng2021secureboost,fan2024secureboost+}.
Initial research in MHFL primarily concentrated on addressing heterogeneity in model architectures. Several methods have been introduced to accommodate clients with different model architectures participating in a federated learning task. These methods typically involve techniques such as knowledge distillation~\cite{hinton2015distilling}, mutual learning and split learning that can handle heterogeneous models.
Knowledge distillation-based MHFL methods, such as FedMD~\cite{li2019fedmd} and FedET~\cite{cho2022heterogeneous}, involve the server aggregating the output logits of different clients' heterogeneous models on a public dataset to construct global logits.
Mutual learning-based MHFL, such as Deep Mutual Learning (DML) \cite{zhang2018deep}, PFML~\cite{yang2021regularized} and FedLoRA \cite{yi2023fedlora}, design a small homogeneous model and a large heterogeneous model in each client. 
Split learning-based MHFL approaches, such as FedClassAvg~\cite{jang2022fedclassavg} and CHFL \cite{liu2022completely}, share a homogeneous classifier to improve model classification while personalizing the local feature extractor.

\subsection{Federated Learning for LLMs}

Parameter-Efficient Fine-Tuning (PEFT) methods \cite{houlsby2019parameter,he2021towards,lester2021power,li2021prefix,hu2021lora} offer a direct solution to the issues of communication overhead and fine-tuning costs in federated learning (FL) for LLMs. A number of studies have built upon PEFT methods in the context of FL for LLMs, including FedPETuning~\cite{zhang2022federated}, Federated Adapter Tuning~\cite{cai2022autofednlp}, Federated Prompt Tuning~\cite{zhao2022reduce}, and FATE-LLM ~\cite{fan2023fate, fan2024fedcollm}. 
These findings indicate that FL clients, especially those with constrained computing and storage resources such as certain devices, can significantly profit from from PEFT approaches. These techniques facilitate the sharing of LLMs across diverse tasks, while necessitating the storage and updating of only a small number of parameters for each task, thereby reducing the overall computational and storage requirements. By leveraging PEFT methods, FL clients can efficiently adapt LLMs to their specific needs while minimizing communication overhead and fine-tuning costs.

\section{The Proposed FedMKT Method}

In this section, we introduce FedMKT, an innovative and parameter-efficient federated mutual knowledge transfer approach for large and small language models. The FedMKT primarily comprises two key modules: \textit{Bidirectional Token Alignment} and  \textit{Selective Mutual Knowledge Transfer}. We will elaborate on these two key modules in Section \ref{sec:token align} and Section \ref{sec:selective knowledge}, respectively after we define the problem we try to address in Section \ref{sec:problem}.

\subsection{Problem Definition}
\label{sec:problem}
We consider the federated learning setting, involving one server that owns an LLM $f_\psi$ parameterized by $\psi$ and $K$ clients that each client $k$ has an SLM $g_{\phi_k}$ parameterized by ${\phi}_k$. Each client owns a local private dataset denoted as $\mathcal{D}_k$ containing $N$ training samples, and all clients and server have access to a shared public dataset $\mathcal{D}_{p}$. 

The server and clients aim to collaboratively enhance the performance of the LLM and SLMs through federated learning without disclosing any private data. We assume that the $K$ clients execute the same text generation task, but they may hold heterogeneous or homogeneous SLM models. The collaboration between clients and the server involves the following sub-procedures: 
\begin{itemize}
    \item Each client $k$ trains its SLM $g_{\phi_k}$ using its private data $\mathcal{D}_k$. The objective is formulated as follows:
     \begin{equation}
    \min_{\phi_1,\phi_2,...,\phi_K} \mathcal{L}_1(\phi_1,\phi_2,...,\phi_K;\{\mathcal{D}_k\}_{k=1}^K) \label{eq:hfl}
    \end{equation}
     
    \item Each client computes the output logits on $\mathcal{D}_{p}$ and securely uploads them to the server. Upon receiving output logits of all clients, the server computes the distillation loss by comparing these client logits with the output logits produced by its own LLM on $\mathcal{D}_{p}$. The objective can be formulated as follows:
    \begin{equation}
        \min_{\psi} \mathcal{L}_2(\psi; \mathcal D_{p}, {\phi_1,\phi_2,...,\phi_K}) \label{eq:mkt-2}
    \end{equation}
    The server aims to transfer knowledge from the clients' SLMs $g_{\phi_k}$ to its owned LLM $f_{\psi}$.

    \item The server dispatches the LLM’s output logits on $\mathcal{D}_{p}$ to all the clients. Subsequently, the clients compute the distillation loss by comparing LLM output logits with SLMs' output logits on $\mathcal{D}_{p}$. The objective can be formulated as follows:
    \begin{equation}
        \min_{\phi_1,\phi_2,...,\phi_K} \mathcal{L}_3(\phi_1,\phi_2,...,\phi_K; \mathcal D_{p}, { \psi}) \label{eq:mkt-3}
    \end{equation}
The clients aim to transfer knowledge from LLM $f_{\psi}$ to enhance their SLMs. 
    
\end{itemize}

We consider the server \textit{semi-honest}, meaning that the server may try to recover the private data of clients from the information it observes.

FedMKT solves the optimization problems formulated in Eq.(\ref{eq:hfl}), Eq.(\ref{eq:mkt-2}), and Eq.(\ref{eq:mkt-3}) in an efficient and privacy-preserving manner. We illustrate the workflow of FedMKT in Figure~\ref{fig:fedmkt} and Appendix~\ref{sec:appendix-workflow}, elaborate on the associated training algorithm in Algorithm \ref{alg:fedmkt}.

\begin{algorithm}[!ht] 
\caption{FedMKT}
\label{alg:fedmkt}
\begin{algorithmic}[1]
\renewcommand{\algorithmicrequire}{\textbf{Input:}}
\renewcommand{\algorithmicensure}{\textbf{Output:}}

\Require  \\
    $K$:  number of clients; \\
    $T$: total number of communication rounds; \\
    $R$: local number of rounds in the server; \\
    $E$: local number of rounds in the client; \\
    $\eta_\omega$: the learning rate of LLM $f_{\psi+\omega}$; \\
    $\eta_\theta$: the learning rate of SLM $g_{\phi_k+\theta_k}$.
    
\Ensure $f_{\psi+\omega}$, $g_{\phi_1+\theta_1}$,$g_{\phi_2+\theta_2}$,...,$g_{\phi_K+\theta_K}$.

\State //$\textbf{ Server side:}$
 \For{$t$ in communication round $T$}
 
        \State  $\{\mathcal{S}_k\}_{k=1}^K$ $\gets \textbf{ClientUpdate1}(t)$.

         \State {Token Alignment} from SLMs to LLM.

        \State $\Tilde{\mathcal{S}}_0 \leftarrow \textbf{DualMinCE}(\mathcal{D}_p, f_{\psi+\omega}, \{\mathcal{S}_k\}_{k=1}^K)$.

        \State \gray{// knowledge transfer based on $\mathcal D_{p}$ and $\Tilde{\mathcal{S}}_0$.}
        \For{each epoch $r \in [R]$}
         \State $\omega^{t, r+1} \gets  \omega^{t, r} - \eta_\omega \bigtriangledown\mathcal{L}_2 $.
         
         \EndFor
        
        \State $\omega^{t+1} = \omega^{t,R} $.

        \State Compute $\mathcal{S}_0 = \{l^i_{0}, p^i_{0}\}_{i=1}^N$ based on $\mathcal{D}_p$.

        \State $\textbf{ClientUpdate2}(t,\mathcal{S}_0)$.

    \EndFor 

        \State
        
        \State $\textbf{ClientUpdate1($t$):}$
        \For{each client $k$ (in parallel)}
            \State \gray{// local fine-tuning based on $\mathcal D_{k}$.}
            \For{each local epoch $e \in [E]$}
            \State $\theta_{k}^{t,e+1} \gets \theta_{k}^{t,e} - \eta_\theta \bigtriangledown \ell_{\text{TA}}$.
            \EndFor
            
            \State Compute $\mathcal{S}_k=\{l^i_{k}, p^i_{k}\}_{i=1}^N$ based on $\mathcal{D}_p$.
      
        \EndFor
         \State Upload $\{\mathcal{S}_k\}_{k=1}^K$ to the server
         
         \State
        \State \textbf{ClientUpdate2($t,\mathcal{S}_0$):}
        \For{each client $k$ (in parallel)}
         
            \State {Token Alignment } from LLM to SLMs.

             \State $\Tilde{\mathcal{S}}_k \leftarrow$ $\textbf{DualMinCE}(\mathcal{D}_p$, $g_{\phi_k+\theta_k}$, $\mathcal{S}_0$).
            \State \gray{// knowledge transfer based on $\mathcal D_{p}$ and $\Tilde{\mathcal{S}}_k$.}
            \For{each local epoch $e \in [E,2E]$}
             \State $\theta^{t,e+1}_k \gets  \theta^{t, e}_k - \eta_\theta \bigtriangledown\mathcal{L}_3 $.
         \EndFor
        
        \State $\theta^{t+1}_k = \theta^{t,2E}_k $.

        \EndFor
        
\end{algorithmic}
\end{algorithm}

\begin{algorithm}[htbp] 
\caption{DualMinCE}
\label{alg:dualmince}
\begin{algorithmic}[1]
\renewcommand{\algorithmicrequire}{\textbf{Input:}}
\renewcommand{\algorithmicensure}{\textbf{Output:}}

\Require  \\
    $\mathcal{D}_p$: the public dataset; \\
    $h$: either the SLM $g_{\phi_k+\theta_k}$ of client $k$ or the LLM $f_{\psi+\omega}$ of the server; \\
    $\mathcal{S}_k=\{(l_k^i, p_k^i)\}_{i=1}^N, k=0 \text{ or } \lceil K \rceil$: loss-logit pairs passed from either the server or clients.
    
\Ensure $\mathcal{S}$.
\State $\Tilde{\mathcal{S}} \leftarrow \{\}$  \gray{$\triangleright$ initialize an empty set of selective knowledge.}
 \For{each $x^i$ in $\mathcal{D}_p$}
      \State $l^i_\text{local} \leftarrow h(x^i)$
       
      \State $  k^{*}  = \begin{cases} \arg\min\limits_{k}{(l^i_k)}, &\text{if } k = \lceil K \rceil \\ 0, &\text{if } k = 0  \end{cases}$

      \State $\Tilde{\mathcal{S}} \leftarrow \Tilde{\mathcal{S}} + (x^i, p^i_{k^{*}})$ $\text{ if }$ $l^i_{k^{*}} < l^i_{\text{local}}$
 \EndFor 
\end{algorithmic}
\end{algorithm}

\subsection{Bidirectional Token Alignment} 
\label{sec:token align}
A significant challenge in aligning output logits distributions lies in the mismatch between tokenizers of different LLM and SLMs, exemplified by Bloom and LLaMa. Consider the sentence, "we utilize the dynamic programming approach to align tokens" as an example. Utilizing the Bloom tokenizer would segment it into the following tokens: ['we', 'utilize', 'the,' 'dynamic,' 'programming,' 'approach,' 'to,' 'align,' 'tokens']. However, if the LLaMa tokenizer were used, the segmentation would be: ['we', 'util', 'ize', 'the', 'dynamic', 'programming', 'approach', 'to', 'align', 'tokens']. 

To tackle this issue, we adopt dynamic programming techniques to promote robust alignment, as evidenced in studies \cite{wan2024knowledge,fu2023specializing}.
Utilizing LLaMa2 and Bloom as illustrative examples, we establish an optimized vocabulary mapping table based on \textit{minimum edit distance(MinED)}. This mapping table identifies the closest Bloom token for each LLaMa2 token (e.g., 'utilize' for 'util'). We then tokenize a sentence using both tokenizers and apply a dynamic programming algorithm to determine the optimal matching path. When multiple LLaMa2 tokens align to a single Bloom token (e.g., 'util' and 'ize' aligning to 'utilize'), we handle them according to the mapping table. 
Please see Appendix \ref{sec:appendix-MinED} for further details.

In FedMKT, a bidirectional token alignment process occurs before knowledge transfer between LLMs and SLMs. One the one hand, when clients transfer knowledge from their SLMs to the server's LLM, the server aligns SLM tokens to LLM tokens. On the other hand, when the server transfers knowledge from its LLM back to clients' SLMs, each client aligns LLM tokens to its SLM tokens.

\subsection{Selective Mutual Knowledge Transfer Between LLM and SLMs}
\label{sec:selective knowledge}

To transfer knowledge between the server and clients efficiently, we leverage LoRA to fine-tune the server's LLM and clients' SLMs. Specifically, 
each client $k$ inserts a small low-rank adapter parameterized by $\theta_k$ into its local SLM. We denote client $k$ local SLM with the added $\theta_k$ as $g_{\phi_k+\theta_k}$. Likewise, the server inserts a small low-rank adapter parameterized by $\omega$ into its LLM $f_{\psi}$. We denote the server's LLM $f_{\psi}$ with the added $\omega$ as $f_{\psi+\omega}$. During the whole federated learning training process, $\theta_k, k=1,...,K$ and $\omega$ are trained, while $\phi_k,k=1,...,K$ and $\psi$ are frozen.

Before transferring knowledge to the server, each client $k$ trains its LoRA adapter $\theta_k$ using its private dataset $\mathcal{D}_k$. Consequently, Eq.(\ref{eq:hfl}) can be reformulated as follows:
\begin{equation}
\begin{aligned}
     \mathcal{L}_1(\theta_1,\theta_2,...,\theta_K;\{\mathcal{D}_k\}_{k=1}^K) \\ =
     \frac{1}{K}\sum_{k=1}^K \mathbb{E}_{(x,y) \sim \mathcal{D}_k} \ell_{\text{TA}}(&g_{\phi_k+\theta_k}(x), y)
\end{aligned}
\end{equation}
where $\ell_{\text{TA}}$ is the task loss for training $\theta_k$ of each client $k$. The original model parameter $\phi_k$ of client $k$'s SLM is frozen during training.

Then, both the server and clients fine-tune their LoRA adapters based on a shared public dataset $\mathcal{D}_p$. We formulate the losses of fine-tuning $f_{\psi+\omega}$ and $g_{\phi_k+\theta_k}$ (denoted as $\mathcal{L}_{\text{FT}}^f$
and $\mathcal{L}_{\text{FT}}^g$) as follows:
\begin{equation}\label{eq:mft-f}
\begin{aligned}
     \mathcal{L}^f_{\text{FT}}(\omega;\mathcal{D}_p) &= \mathbb{E}_{(x,y) \sim \mathcal{D}_p}\ell_{\text{CE}}(f_{\psi+\omega}(x), y)
\end{aligned}
\end{equation}
\begin{equation}\label{eq:mft-g}
\begin{aligned}
     \mathcal{L}^g_{\text{FT}}(\theta_k;\mathcal{D}_p)&= \mathbb{E}_{(x,y) \sim \mathcal{D}_p} \ell_{\text{CE}}(g_{\phi_k+\theta_k}(x),y)
\end{aligned}
\end{equation}
where $\ell_{\text{CE}}$ represents the cross-entropy loss, and the model parameters $\psi$ and $\phi_k$ are frozen during fine-tuning. 

Next, the server and clients conduct selective knowledge transfer to each other. The motivation for applying selective knowledge transfer is that some clients' knowledge may adversely affect the performance of LLM on the server and vice versa in a heterogeneous environment. Therefore, it is critical to guarantee that the knowledge exchanged between the server and clients is positive to the performance of LLM and SLMs. To achieve this goal, we propose a selective knowledge transfer strategy on both the server and client sides, termed \textit{DualMinCE}.

DualMinCE aims to select knowledge that is positive to the performance of the server's LLM from clients and vice versa. Specifically,  when knowledge needs to be transferred from SLMs to the LLM, each client $k$ computes a knowledge set $\mathcal{S}_k=\{l^i_{k}, p^i_{k}\}_{i=1}^N$ consisting of loss-logit pairs through its local model based on the public dataset $\mathcal{D}_p$.  Then, all $K$ clients send their $\{\mathcal{S}_k\}_{k=1}^K$ to the server. By leveraging DualMinCE (see Algorithm \ref{alg:dualmince} for detail), the server picks a logit $p_{k^*}^i$ with the smallest loss from $\{l^i_{k}, p^i_{k}\}_{k=1}^K$ and adds $p_{k^*}^i$ to a selective knowledge set $\Tilde{\mathcal{S}}_0$ if the loss $l_{k^*}^i$ of $p_{k^*}^i$  is smaller than the loss $l_{\text{local}}^i$ computed through the server's local LLM based on $x^i$  for each $x^i$ in $\mathcal{D}_p$. 

Next, the server leverages the knowledge distillation loss, denoted as $\mathcal{L}_{\text{KD}}^f$, to fine-tune $f_{\psi+\omega}$:

\begin{equation}\label{eq:mkd-f}
\begin{aligned}
     \mathcal{L}^f_{\text{KD}}(\omega;\Tilde{\mathcal{S}}_0) = &\mathbb{E}_{(x,p) \sim \Tilde{\mathcal{S}}_0} \ell_{\text{CE}}(f_{\psi+\omega}(x), p)
\end{aligned}
\end{equation}

Likewise, each client $k$ leverages DualMinCE to form its selective knowledge set $\Tilde{\mathcal{S}}_k$ from the knowledge $\mathcal{S}_0$ sent from the server. Each client $k$ leverages the following knowledge distillation loss to fine-tune its local model $g_{\phi_k+\theta_k}$:
\begin{equation}\label{eq:mkd-g}
\begin{aligned}
     \mathcal{L}^g_{\text{KD}}(\theta_k;\Tilde{\mathcal{S}}_k) = &\mathbb{E}_{(x,p) \sim \Tilde{\mathcal{S}}_k} \ell_{\text{CE}}(g_{\phi_k+\theta_k}(x),p)
\end{aligned}
\end{equation}

Combining Eq.(\ref{eq:mft-f}) and Eq.(\ref{eq:mkd-f}), we reformulate the knowledge transfer from SLMs to LLM conducted on the server to enhance LLM as follows:
\begin{equation}\label{eq:fedmhllm-llm}
\begin{aligned}
   \mathcal{L}_2 &= \lambda \mathcal{L}_{\text{FT}}^f + (1 - \lambda )\mathcal{L}_{\text{KD}}^f\\
\end{aligned}
\end{equation}

Combining Eq.(\ref{eq:mft-g}) and Eq.(\ref{eq:mkd-g}), we reformulate the knowledge transfer from LLM to SLMs conducted on the clients to enhance SLMs as follows:
\begin{equation}\label{eq:fedmhllm-slm}
\begin{aligned}
    \mathcal{L}_3 & =\frac{1}{K}\sum_{k=1}^K (\lambda  \mathcal{L}_{\text{FT}}^g + (1 - \lambda) \mathcal{L}_{\text{KD}}^g)
\end{aligned}
\end{equation}
where $\lambda$ is the hyperparameter that regulates the significance of mutual knowledge transfer.

\begin{table*}[!ht]
\centering
\setlength{\tabcolsep}{3.2pt}
\begin{tabular}{c|c|c|c|c|c}
\hline
\hline
\multicolumn{1}{c}{\textbf{Setting}} & \multicolumn{1}{c}{\textbf{Server}} & \multicolumn{1}{c}{\textbf{Client-1}} &    \multicolumn{1}{c} {\textbf{Client-2}} & \multicolumn{1}{c} {\textbf{Client-3}} & \multicolumn{1}{c} {\textbf{Client-4}}\\
\hline

\textbf{Heterogeneous}& LLaMa2-
7B& GPT-2-xlarge(1.5B)& OPT-1.3B & Bloom-1.1B & LLaMa2-1.3B\\
\hline
\textbf{Homogeneous}& LLaMa2-
7B& LLaMa2-1.3B& LLaMa2-1.3B& LLaMa2-1.3B& LLaMa2-1.3B\\
\hline
\textbf{ Homogeneous}& LLaMa2-
7B& OPT-1.3B & OPT-1.3B & OPT-1.3B & OPT-1.3B \\
\hline

\textbf{One-to-One}& LLaMa2-
7B& -& -& -& LLaMa2-1.3B\\
\hline

\textbf{One-to-One}& LLaMa2-
7B& -& OPT-1.3B & -& -\\

\hline
\hline
\end{tabular}
\caption{The five different settings we utilize to evaluate FedMKT.}
\label{tab: model setting}
\end{table*}

\begin{table*}[!t]
\centering
\setlength{\tabcolsep}{3.2pt}
\begin{tabular}{c|c|c|c|c|c|c}
\hline
\hline
\multicolumn{1}{c}{\textbf{Task}} & \multicolumn{1}{c}{\textbf{Method}} & \multicolumn{1}{c}{\textbf{GPT-2-xlarge}}   & \multicolumn{1}{c}{\textbf{OPT-1.3B}} &\multicolumn{1}{c}{\textbf{Bloom-1.1B}} 
 &\multicolumn{1}{c}{\textbf{LLaMa2-1.3B}}  &\multicolumn{1}{c}{\textbf{LLaMa2-7B}}\\
\hline

\multirow{4}*{RTE}& Centralized & -& -& -& -& 85.9\\

\cline{2-7}
~& Zero-Shot& 52.4& 52.7& 52.7 & 49.8&63.2\\

\cline{2-7}
~& Standalone & 65.7& 62.5& 58.1& 55.6&-\\ 

\cline{2-7}
~        & \textbf{FedMKT}& \textbf{70.4}& \textbf{65.7}& \textbf{61.7}& \textbf{58.8} &82.3\\

\hline

\multirow{4}*{WIC} & Centralized  & -& -& -& -& 70.4\\
\cline{2-7}

~ & Zero-Shot& 49.8& 50.8& 50 & 50&50.3\\
\cline{2-7}

~ & Standalone & 59.3& 52.2& 59.1& 50.6&-\\ 
\cline{2-7}

~ & \textbf{FedMKT}& \textbf{63.2}& \textbf{62.2}& \textbf{61.1}& \textbf{51.9}&61.3\\ 

\hline

\multirow{4} *{BoolQ}& Centralized  & -& -& -& -& 87.6\\

\cline{2-7}
~ &  Zero-Shot& 61.3& 58.4& 59.0 & 61.0&70.1\\

\cline{2-7}
~         & Standalone & 71.1& 74.1& 69.7& 69.9&-\\ 

\cline{2-7}
~        & \textbf{FedMKT}& \textbf{75.1}& \textbf{76.8}& \textbf{71.4}& \textbf{75.1}&85.0\\ 

\hline

\multirow{4}*{CQA}& Centralized  & -& -& -& -& 69.5\\

\cline{2-7}
~ &  Zero-Shot& 36.7& 41.9& 33.8 & 30.1&39.5\\

\cline{2-7}
~ & Standalone & 56.0& 58.6& 44.7& 56.7&-\\ 

\cline{2-7}
~ & \textbf{FedMKT}& \textbf{58.3}& \textbf{60.5}& \textbf{50.8}& \textbf{57.0}&71.8\\ 

\hline

\multirow{4}*{ARC-E}& Centralized & -& -& -& -& 76.9\\

\cline{2-7}
~ & Zero-Shot& 58.3& 57.0& 51.5 & 53.1&69.3\\

\cline{2-7}
~ & Standalone & 59.3& 57.9& 56.9& 60.4&-\\ 

\cline{2-7}
~        & \textbf{FedMKT}& \textbf{59.8}& \textbf{59.6}& \textbf{57.5}& \textbf{60.8}&76.1\\ 

\hline

\multirow{4}*{ARC-C} & Centralized  & -& -& -& -& 48.9\\

\cline{2-7}
~ &  Zero-Shot& 25.0& 23.4& 23.6 & 26.7&40.0\\

\cline{2-7}
~ & Standalone & 28.2& 28.4& 24.9& 28.5&-\\ 

\cline{2-7}
~ & \textbf{FedMKT}& \textbf{30.2}& \textbf{29.4}& \textbf{26.6}& \textbf{30.0}&44.7\\ 

\hline

\multirow{4}*{S-NI} & Centralized & -& -& -& -& 49.3\\

\cline{2-7}
~ &  Zero-Shot& 5.0& 5.2& 5.1 & 5.8&12.0\\

\cline{2-7}
~ & Standalone & 27.9& 26.1& 10.6&33.4&-\\

\cline{2-7}
~        & \textbf{FedMKT}& \textbf{34.2}& \textbf{36.0}& \textbf{15.1}& \textbf{37.3}&41.4\\ 

\hline

\multirow{4}*{DialogSum} & Centralized  & -& -& -& -& 27.7\\

\cline{2-7}
~ &  Zero-Shot& 5.4& 6.4& 4.9 & 5.7&8.5\\

\cline{2-7}
~ & Standalone & 22.3& 19.8& 13.2& 21.4&-\\ 

\cline{2-7}
~ & \textbf{FedMKT}&\textbf{23.2}& \textbf{20.9}& \textbf{14.9}& \textbf{21.6}&24.2\\ 

\hline
\hline

\end{tabular}
\caption{
Method Performance Comparison in \textbf{the Heterogeneous setting}. We evaluate FedMKT with 8 different tasks. In all the 8 tasks, the server is deployed with a LLaMa2-7B model, and the 4 clients are deployed with a GPT-2-xlarge, a OPT-1.3B, a Bloom-1.1B, and a LLaMa2-1.3B, respectively. The '-' indicates a method does not apply to the corresponding participant (either the server or the client).}
\label{tab:Heterogeneous}
\end{table*}

\section{Experiments}

\subsection{Setup}
We set up a federated learning scenario involving four clients and one server to evaluate the FedMKT using various publicly available LLMs and SLMs.

\textbf{Models}. We evaluate FedMKT on one LLM (LLaMa2-7B \cite{touvron2023llama}) in the server, four SLMs in the clients including GPT-2-xlarge (1.5B) \cite{radford2019language}, OPT-1.3B \cite{zhang2022opt}, Bloom-1.1B \cite{scao2022bloom} and LLaMa2-1.3B \cite{xia2023sheared}.  In our experiments, we evaluate our framework in three distinct scenarios: \textbf{Heterogeneous}, \textbf{Homogeneous} and \textbf{One-to-One}.
Table \ref{tab: model setting} details the setup for the LLM and SLMs in different settings.

\textbf{Datasets}. We evaluate FedMKT on 6 QA datasets and 2 instruction-following datasets. Specifically, for QA tasks, we use RTE \cite{wang2019superglue}, WTC~\cite{wang2019superglue},  BoolQ~\cite{clark2019boolq}, CQA ~\cite{talmor2018commonsenseqa}, ARC-E and ARC-C~\cite{clark2018think} to evaluate FedMKT. As for instruction-following tasks, we evaluate FedMKT on S-NI~\cite{wang2022benchmarking},
DialogSum~\cite{chen2021dialogsum}. 

\textbf{Baselines}. 
We conduct a comparative analysis of FedMKT against the following baselines:
\begin{itemize}
    \item Centralized, in which the server's LLM is fine-tuned locally using the datasets combining private datasets of involved clients and the public dataset. In the One-to-One setting, the data of one client and the public data are used to fine-tune the server's LLM, whereas in other settings, the data of all four clients and the public data are utilized to fine-tune the LLM; 
    \item Zero-Shot, representing the zero-shot capabilities of LLM or SLMs (without fine-tuning); 
    \item Standalone, where each client fine-tunes its own SLM independently using its private dataset;
    \item FedAvg\cite{mcmahan2017communication}, representing the standard federated averaging algorithm. FedAvg is only used in homogeneous settings because it requires all clients' models have the same architecture.
    \item LLM2SLM, representing FedMKT involving one server with an LLM and one client with an SLM. The LLM is not updated and is used to transfer knowledge to SLM. LLM2SLM is only used in the One-to-One setting.
\end{itemize}

\textbf{Evaluation Metrics}. 
For the QA datasets, we primarily use \textbf{Accuracy} as the metric for evaluation, whereas for the instruction-following datasets, we primarily rely on \textbf{Rouge-L}. It's worth noting that in our experiments, all methods across the three scenarios undergo zero-shot evaluation, and we utilize the \textit{lm-evaluation-harness} package\cite{eval-harness} for evaluation purposes.

\subsection{Evaluation on Heterogeneous Setting}

In the Heterogeneous setting, the server is deployed with a LLaMa2-7B model, and the 4 clients are deployed with a GPT-2-xlarge, a OPT-1.3B, a Bloom-1.1B, and a LLaMa2-1.3B, respectively. Table \ref{tab:Heterogeneous} reports the performance comparisons of FedMKT against baselines on 8 tasks. 

Tables \ref{tab:Heterogeneous} show that FedMKT performs superior over Zero-Shot and Standalone on all clients' SLMs. Take the RTE dataset as an example, 
FedMKT outperforms Zero-Shot by 34\% and Standalone by 7\% in relative terms on the GPT-2-xlarge SLM; FedMKT surpasses Zero-Shot by 25\% and Standalone by 5\% on the OPT-1.3B SLM; FedMKT-SLM achieves a 17\% improvement over Zero-Shot and a 6\% improvement over Standalone on the Bloom-1.1B SLM; FedMKT-SLM outperforms Zero-Shot by 18\% and Standalone by 6\% on the LLaMa2-1.3B SLM.
These empirical results demonstrate that, by leveraging FedMKT, SLMs are able to effectively leverage the knowledge transferred from the LLM, leading to enhanced model capabilities. 

Table \ref{tab:Heterogeneous} also shows that FedMKT outperforms Zero-Shot and Centralized on the LLaMa2-7B of the server. For instance, on the RTE QA dataset, FedMKT outperforms Zero-Shot by 30\% and achieves a performance level that is nearly on par with Centralized, reaching approximately 96\% of its fine-tuning performance. 
This significant achievement signifies that FedMKT effectively facilitates the acquisition of knowledge from all clients by the server. 

\subsection{Evaluation on Homogeneous Setting}

We conduct experiments with two Homogeneous settings, as shown in Table \ref{tab: model setting}. The first setting (denoted as S1) involves one server-side LLaMa2-7B and four client-side LLaMa2-1.3B. The second setting (denoted as S2) involves one server-side LLaMa2-7B and four client-side OPT-1.3B. 

Table \ref{tab:Homogeneous} reports the performance comparisons of FedMKT against baselines in the two Homogeneous settings. The top sub-table and the bottom sub-table compare the performance of FedMKT against baselines on the server's LLM and clients' SLMs, respectively.

The top sub-table of Table \ref{tab:Homogeneous} shows that FedMKT significantly outperforms Zero-Shot on the server's LLM (i.e., LLaMa2-7B) in the two Homogeneous settings. It also shows that FedMKT achieves comparable performance of the Centralized scenario, in which the server' LLM is fine-tuned using all clients' data and the public data combined.

The bottom sub-table of Table \ref{tab:Homogeneous} shows that FedMTK performs better than the Zero-Shot, Standalone, and FedAvg due to the assistance of the server's LLM. For example, in the CQA dataset, FedMKT outperforms FedAvg by 4\% in relative terms on the LLaMa2-1.3 SLM and by 5\% on the OPT-1.3B SLM, respectively.

\begin{table}[!ht]
\centering
\setlength{\tabcolsep}{2.8pt}

\begin{tabular}{c|c|c|c}
\hline
 \hline
\multirow{2}{*}{\shortstack{\textbf{Task}}} & \multirow{2}{*}{\shortstack{{\textbf{Method}}}} & 
\multirow{2}{*}{\shortstack{S1: Server \\\textbf{LLaMa2-7B}}}  &
\multirow{2}{*}{\shortstack{S2: Server \\ \textbf{LLaMa2-7B}}}\\
~ & ~ & ~ & ~ \\  

\hline

\multirow{3}*{CQA}& Zero-Shot&39.5&39.5\\
\cline{2-4}
~& Centralized&69.5&69.5\\
 \cline{2-4}
~& \textbf{FedMKT}& 68.8&71.3\\ 

 \hline

\multirow{3}*{ARC-C}& Zero-Shot&40.0&40.0\\

\cline{2-4}
~& Centralized&49.4&49.4\\
 \cline{2-4}
~& \textbf{FedMKT}& 46.2&46.2\\ 

 \hline

\multirow{3}*{ARC-E}& Zero-Shot&69.3&69.3\\

\cline{2-4}
~& Centralized&75.5&75.5\\
 \cline{2-4}
~& \textbf{FedMKT}& 74.9&74.8\\
 \hline

\end{tabular}

\begin{tabular}{c|c|c|c}
\hline
\multirow{2}{*}{\shortstack{\textbf{Task}}} & \multirow{2}{*}{\shortstack{{\textbf{Method}}}} & 
\multirow{2}{*}{\shortstack{S1: Clients \\ \textbf{LLaMa2-1.3B}}}  &
\multirow{2}{*}{\shortstack{S2: Clients \\ \textbf{OPT-1.3B}}}\\
~ & ~ & ~ & ~ \\  

\hline

\multirow{4}*{CQA}& Zero-Shot&30.1&41.9 \\

\cline{2-4}
~& Standalone&56.4&58.1\\
\cline{2-4}
~& FedAvg&56.4&58.6\\
 \cline{2-4}
~& \textbf{FedMKT}& \textbf{58.6}&\textbf{61.5}\\

 \hline
 
\multirow{4}*{ARC-C}& Zero-Shot&26.7&23.4\\

\cline{2-4}
~& Standalone&30.4&28.5\\

\cline{2-4}
~& FedAvg&29.7&28.6\\
 \cline{2-4}
~& \textbf{FedMKT}& \textbf{31.7}&\textbf{29.9}\\

 \hline
 
\multirow{4}*{ARC-E}& Zero-Shot&53.1&57.0\\

\cline{2-4}
~& Standalone&60.3&57.9\\

\cline{2-4}
~& FedAvg&60.6&58.8\\
 \cline{2-4}
~& \textbf{FedMKT}& \textbf{61.7}&\textbf{60.1}\\

 \hline
 \hline

\end{tabular}

\caption{
Method Performance Comparison in \textbf{Homogeneous settings}. We evaluate FedMKT using two homogeneous settings. The first setting (denoted as S1) involves one server-side LLaMa2-7B LLM and four client-side LLaMa2-1.3B SLMs, while the second setting (denoted as S2) involves one server-side LLaMa2-7B LLM and four client-side OPT-1.3B SLMs. \textit{The top and bottom sub-tables compare the performance of FedMKT against baselines on the server's LLM and clients' SLMs, respectively}. The results reported in the bottom sub-table are the average of all clients.}
\label{tab:Homogeneous}
\end{table}

\subsection{Evaluation on One-to-One Setting}

We evaluate FedMKT using two One-to-One settings. The first setting (denoted as S1) involves one server-side LLaMa2-7B LLM and one client-side LLaMa2-1.3B SLM, while the second setting (denoted as S2) involves one server-side LLaMa2-7B LLM and one client-side OPT-1.3B SLM.

Table \ref{tab:one-to-one} reports the performance comparisons of FedMKT against baselines in the two One-to-One settings. The top and bottom sub-tables compare the performance of FedMKT against baselines on the server's LLM and clients' SLMs, respectively.

The top sub-table of Table \ref{tab:one-to-one} shows that FedMKT notably surpasses Zero-Shot and rivals Centralized on the performance of the server's LLM. The bottom sub-table of Table \ref{tab:one-to-one} shows that FedMTK achieves superior SLM performance over Zero-Shot, Standalone, and LLM2SLM due to the assistance of LLM. These empirical results demonstrate the effectiveness of FedMKT in transferring knowledge between the LLM and SLMs.

\begin{table}[!t]
\centering
\setlength{\tabcolsep}{2.8pt}

\begin{tabular}{c|c|c|c}
\hline
 \hline
\multirow{2}{*}{\shortstack{\textbf{Task}}} & \multirow{2}{*}{\shortstack{{\textbf{Method}}}} & 
\multirow{2}{*}{\shortstack{S1: Server \\\textbf{LLaMa2-7B}}}  &
\multirow{2}{*}{\shortstack{S2: Server \\ \textbf{LLaMa2-7B}}}\\
~ & ~ & ~ & ~ \\

\hline

\multirow{3}*{CQA}& Zero-Shot&39.5&39.5\\ 
\cline{2-4}
~& Centralized&69.0&68.3\\ 
\cline{2-4}
~& \textbf{FedMKT}&69.0&71.0\\ 

\hline

\multirow{3}*{ARC-C}& Zero-Shot&40.0&40.0\\ 
\cline{2-4}
~& Centralized&45.9&48.6\\ 
\cline{2-4}
~& \textbf{FedMKT}&45.9&45.8\\ 

\hline

\multirow{3}*{ARC-E}& Zero-Shot&69.3&69.3\\ 
\cline{2-4}
~& Centralized&74.4&73.6\\ 
\cline{2-4}
~& \textbf{FedMKT}&74.8& 74.8\\ 

\hline

\end{tabular}

\begin{tabular}{c|c|c|c}
\hline
\multirow{2}{*}{\shortstack{\textbf{Task}}} & \multirow{2}{*}{\shortstack{{\textbf{Method}}}} & 
\multirow{2}{*}{\shortstack{S1: Clients \\ \textbf{LLaMa2-1.3B}}}  &
\multirow{2}{*}{\shortstack{S2: Clients \\ \textbf{OPT-1.3B}}}\\
~ & ~ & ~ & ~ \\  

\hline

\multirow{4}*{CQA}& Zero-Shot&30.1&41.9 \\

\cline{2-4}
~& Standalone&56.7&58.6\\ 

\cline{2-4}
~& LLM2SLM&56.76&59.1\\ 

\cline{2-4}
~& \textbf{FedMKT}&\textbf{56.84} &\textbf{60.7} \\

\hline

\multirow{4}*{ARC-C}& Zero-Shot&26.7&23.4\\

\cline{2-4}
~& Standalone&30.3&28.8\\ 
\cline{2-4}
~& LLM2SLM&30.1&29.6\\ 

\cline{2-4}
~& \textbf{FedMKT}&\textbf{30.8}&\textbf{30.4}\\

\hline

\multirow{4}*{ARC-E}& Zero-Shot&53.1&57.0\\

\cline{2-4}
~& Standalone&57.0&57.9\\
\cline{2-4}
~& LLM2SLM&60.7&58.4\\ 

\cline{2-4}
~& \textbf{FedMKT}&\textbf{60.8}& \textbf{58.5}\\

\hline
\hline

\end{tabular}

\caption{
Method Performance Comparison in \textbf{One-to-One settings}. We evaluate FedMKT using two one-to-one settings. The first setting (denoted as S1) involves one server-side LLaMa2-7B LLM and one client-side LLaMa2-1.3B SLM, while the second setting (denoted as S2) involves one server-side LLaMa2-7B LLM and one client-side OPT-1.3B SLM. \textit{The top and bottom sub-tables compare the performance of FedMKT against baselines on the server's LLM and a client's SLM, respectively}. }

\label{tab:one-to-one}
\end{table}

\section{Conclusions}

In this study, we have presented FedMKT, a parameter-efficient federated mutual knowledge transfer framework tailored for LLMs and SLMs. FedMKT bridges the gap between the server-side LLMs and clients' SLMs, enabling selective mutual knowledge transfer while preserving data privacy. 
Through extensive experiments across three distinct scenarios,  we have demonstrated that FedMKT simultaneously boosts the performance of both LLMs and SLMs.

\section*{Limitations}
In this study, we transfer knowledge between the server and clients using logits of a public dataset, motivated by efficiency and privacy considerations. Although empirical evidence suggests that sharing logits of public datasets between the server and clients is more privacy-preserving than sharing model gradients or parameters \cite{li2019fedmd,cho2022heterogeneous}, there is no theoretical guarantee that this approach does not compromise the privacy of clients' sensitive data. This issue warrants further investigation. 
Additionally, the study does not address the potential presence of spammer~\cite{zhu2012discovering} among clients, which could negatively impact the performance of LLM in the server and the SLMs of other clients. Detecting spammers in the FedMKT setting is identified as an important research direction. 
Furthermore, our study is limited by computational and storage constraints, which preclude the exploration of larger language models. This highlights the inherent trade-off between utility and efficiency. Our future research aims to investigate and optimize this trade-off.


\bibliography{ref}

\begin{thebibliography}{45}
\providecommand{\natexlab}[1]{#1}

\bibitem[{Cai et~al.(2022)Cai, Wu, Wang, Lin, and Xu}]{cai2022autofednlp}
Dongqi Cai, Yaozong Wu, Shangguang Wang, Felix~Xiaozhu Lin, and Mengwei Xu. 2022.
\newblock Autofednlp: An efficient fednlp framework.
\newblock \emph{arXiv preprint arXiv:2205.10162}.

\bibitem[{Chen et~al.(2021)Chen, Liu, Chen, and Zhang}]{chen2021dialogsum}
Yulong Chen, Yang Liu, Liang Chen, and Yue Zhang. 2021.
\newblock Dialogsum: A real-life scenario dialogue summarization dataset.
\newblock \emph{arXiv preprint arXiv:2105.06762}.

\bibitem[{Cheng et~al.(2021)Cheng, Fan, Jin, Liu, Chen, Papadopoulos, and Yang}]{cheng2021secureboost}
Kewei Cheng, Tao Fan, Yilun Jin, Yang Liu, Tianjian Chen, Dimitrios Papadopoulos, and Qiang Yang. 2021.
\newblock Secureboost: A lossless federated learning framework.
\newblock \emph{IEEE Intelligent Systems}, 36(6):87--98.

\bibitem[{Cho et~al.(2022)Cho, Manoel, Joshi, Sim, and Dimitriadis}]{cho2022heterogeneous}
Yae~Jee Cho, Andre Manoel, Gauri Joshi, Robert Sim, and Dimitrios Dimitriadis. 2022.
\newblock Heterogeneous ensemble knowledge transfer for training large models in federated learning.
\newblock \emph{arXiv preprint arXiv:2204.12703}.

\bibitem[{Clark et~al.(2019)Clark, Lee, Chang, Kwiatkowski, Collins, and Toutanova}]{clark2019boolq}
Christopher Clark, Kenton Lee, Ming-Wei Chang, Tom Kwiatkowski, Michael Collins, and Kristina Toutanova. 2019.
\newblock Boolq: Exploring the surprising difficulty of natural yes/no questions.
\newblock \emph{arXiv preprint arXiv:1905.10044}.

\bibitem[{Clark et~al.(2018)Clark, Cowhey, Etzioni, Khot, Sabharwal, Schoenick, and Tafjord}]{clark2018think}
Peter Clark, Isaac Cowhey, Oren Etzioni, Tushar Khot, Ashish Sabharwal, Carissa Schoenick, and Oyvind Tafjord. 2018.
\newblock Think you have solved question answering? try arc, the ai2 reasoning challenge.
\newblock \emph{arXiv preprint arXiv:1803.05457}.

\bibitem[{Fan et~al.(2024{\natexlab{a}})Fan, Chen, Ma, Kang, Fan, and Yang}]{fan2024secureboost+}
Tao Fan, Weijing Chen, Guoqiang Ma, Yan Kang, Lixin Fan, and Qiang Yang. 2024{\natexlab{a}}.
\newblock Secureboost+: Large scale and high-performance vertical federated gradient boosting decision tree.
\newblock In \emph{Pacific-Asia Conference on Knowledge Discovery and Data Mining}, pages 237--249. Springer.

\bibitem[{Fan et~al.(2024{\natexlab{b}})Fan, Kang, Chen, Gu, Song, Fan, Chen, and Yang}]{fan2024pdss}
Tao Fan, Yan Kang, Weijing Chen, Hanlin Gu, Yuanfeng Song, Lixin Fan, Kai Chen, and Qiang Yang. 2024{\natexlab{b}}.
\newblock Pdss: A privacy-preserving framework for step-by-step distillation of large language models.
\newblock \emph{arXiv preprint arXiv:2406.12403}.

\bibitem[{Fan et~al.(2023)Fan, Kang, Ma, Chen, Wei, Fan, and Yang}]{fan2023fate}
Tao Fan, Yan Kang, Guoqiang Ma, Weijing Chen, Wenbin Wei, Lixin Fan, and Qiang Yang. 2023.
\newblock Fate-llm: A industrial grade federated learning framework for large language models.
\newblock \emph{arXiv preprint arXiv:2310.10049}.

\bibitem[{Fan et~al.(2024{\natexlab{c}})Fan, Kang, Ma, Fan, Chen, and Yang}]{fan2024fedcollm}
Tao Fan, Yan Kang, Guoqiang Ma, Lixin Fan, Kai Chen, and Qiang Yang. 2024{\natexlab{c}}.
\newblock Fedcollm: A parameter-efficient federated co-tuning framework for large and small language models.
\newblock \emph{arXiv preprint arXiv:2411.11707}.

\bibitem[{Fu et~al.(2023)Fu, Peng, Ou, Sabharwal, and Khot}]{fu2023specializing}
Yao Fu, Hao Peng, Litu Ou, Ashish Sabharwal, and Tushar Khot. 2023.
\newblock Specializing smaller language models towards multi-step reasoning.
\newblock In \emph{International Conference on Machine Learning}, pages 10421--10430. PMLR.

\bibitem[{Gao et~al.(2023)Gao, Tow, Abbasi, Biderman, Black, DiPofi, Foster, Golding, Hsu, Le~Noac'h, Li, McDonell, Muennighoff, Ociepa, Phang, Reynolds, Schoelkopf, Skowron, Sutawika, Tang, Thite, Wang, Wang, and Zou}]{eval-harness}
Leo Gao, Jonathan Tow, Baber Abbasi, Stella Biderman, Sid Black, Anthony DiPofi, Charles Foster, Laurence Golding, Jeffrey Hsu, Alain Le~Noac'h, Haonan Li, Kyle McDonell, Niklas Muennighoff, Chris Ociepa, Jason Phang, Laria Reynolds, Hailey Schoelkopf, Aviya Skowron, Lintang Sutawika, Eric Tang, Anish Thite, Ben Wang, Kevin Wang, and Andy Zou. 2023.
\newblock \href {https://doi.org/10.5281/zenodo.10256836} {A framework for few-shot language model evaluation}.

\bibitem[{Gu et~al.(2023)Gu, Dong, Wei, and Huang}]{gu2023minillm}
Yuxian Gu, Li~Dong, Furu Wei, and Minlie Huang. 2023.
\newblock Minillm: Knowledge distillation of large language models.
\newblock In \emph{The Twelfth International Conference on Learning Representations}.

\bibitem[{He et~al.(2021)He, Zhou, Ma, Berg-Kirkpatrick, and Neubig}]{he2021towards}
Junxian He, Chunting Zhou, Xuezhe Ma, Taylor Berg-Kirkpatrick, and Graham Neubig. 2021.
\newblock Towards a unified view of parameter-efficient transfer learning.
\newblock \emph{arXiv preprint arXiv:2110.04366}.

\bibitem[{Hinton et~al.(2015)Hinton, Vinyals, and Dean}]{hinton2015distilling}
Geoffrey Hinton, Oriol Vinyals, and Jeff Dean. 2015.
\newblock Distilling the knowledge in a neural network.
\newblock \emph{arXiv preprint arXiv:1503.02531}.

\bibitem[{Houlsby et~al.(2019)Houlsby, Giurgiu, Jastrzebski, Morrone, De~Laroussilhe, Gesmundo, Attariyan, and Gelly}]{houlsby2019parameter}
Neil Houlsby, Andrei Giurgiu, Stanislaw Jastrzebski, Bruna Morrone, Quentin De~Laroussilhe, Andrea Gesmundo, Mona Attariyan, and Sylvain Gelly. 2019.
\newblock Parameter-efficient transfer learning for nlp.
\newblock In \emph{International Conference on Machine Learning}, pages 2790--2799. PMLR.

\bibitem[{Hu et~al.(2021)Hu, Shen, Wallis, Allen-Zhu, Li, Wang, Wang, and Chen}]{hu2021lora}
Edward~J Hu, Yelong Shen, Phillip Wallis, Zeyuan Allen-Zhu, Yuanzhi Li, Shean Wang, Lu~Wang, and Weizhu Chen. 2021.
\newblock Lora: Low-rank adaptation of large language models.
\newblock \emph{arXiv preprint arXiv:2106.09685}.

\bibitem[{Jang et~al.(2022)Jang, Ha, Jung, and Yoon}]{jang2022fedclassavg}
Jaehee Jang, Heoneok Ha, Dahuin Jung, and Sungroh Yoon. 2022.
\newblock Fedclassavg: Local representation learning for personalized federated learning on heterogeneous neural networks.
\newblock In \emph{Proceedings of the 51st International Conference on Parallel Processing}, pages 1--10.

\bibitem[{Kang et~al.(2023)Kang, Fan, Gu, Fan, and Yang}]{kang2023grounding}
Yan Kang, Tao Fan, Hanlin Gu, Lixin Fan, and Qiang Yang. 2023.
\newblock Grounding foundation models through federated transfer learning: A general framework.
\newblock \emph{arXiv preprint arXiv:2311.17431}.

\bibitem[{Lester et~al.(2021)Lester, Al-Rfou, and Constant}]{lester2021power}
Brian Lester, Rami Al-Rfou, and Noah Constant. 2021.
\newblock The power of scale for parameter-efficient prompt tuning.
\newblock \emph{arXiv preprint arXiv:2104.08691}.

\bibitem[{Lhoest et~al.(2021)Lhoest, Villanova~del Moral, Jernite, Thakur, von Platen, Patil, Chaumond, Drame, Plu, Tunstall, Davison, {\v{S}}a{\v{s}}ko, Chhablani, Malik, Brandeis, Le~Scao, Sanh, Xu, Patry, McMillan-Major, Schmid, Gugger, Delangue, Matussi{\`e}re, Debut, Bekman, Cistac, Goehringer, Mustar, Lagunas, Rush, and Wolf}]{lhoest-etal-2021-datasets}
Quentin Lhoest, Albert Villanova~del Moral, Yacine Jernite, Abhishek Thakur, Patrick von Platen, Suraj Patil, Julien Chaumond, Mariama Drame, Julien Plu, Lewis Tunstall, Joe Davison, Mario {\v{S}}a{\v{s}}ko, Gunjan Chhablani, Bhavitvya Malik, Simon Brandeis, Teven Le~Scao, Victor Sanh, Canwen Xu, Nicolas Patry, Angelina McMillan-Major, Philipp Schmid, Sylvain Gugger, Cl{\'e}ment Delangue, Th{\'e}o Matussi{\`e}re, Lysandre Debut, Stas Bekman, Pierric Cistac, Thibault Goehringer, Victor Mustar, Fran{\c{c}}ois Lagunas, Alexander Rush, and Thomas Wolf. 2021.
\newblock \href {https://arxiv.org/abs/2109.02846} {Datasets: A community library for natural language processing}.
\newblock In \emph{Proceedings of the 2021 Conference on Empirical Methods in Natural Language Processing: System Demonstrations}, pages 175--184, Online and Punta Cana, Dominican Republic. Association for Computational Linguistics.

\bibitem[{Li and Wang(2019)}]{li2019fedmd}
Daliang Li and Junpu Wang. 2019.
\newblock Fedmd: Heterogenous federated learning via model distillation.
\newblock \emph{arXiv preprint arXiv:1910.03581}.

\bibitem[{Li and Liang(2021)}]{li2021prefix}
Xiang~Lisa Li and Percy Liang. 2021.
\newblock Prefix-tuning: Optimizing continuous prompts for generation.
\newblock \emph{arXiv preprint arXiv:2101.00190}.

\bibitem[{Liu et~al.(2022)Liu, Yang, Cai, Ding, and Lu}]{liu2022completely}
Chang Liu, Yuwen Yang, Xun Cai, Yue Ding, and Hongtao Lu. 2022.
\newblock Completely heterogeneous federated learning.
\newblock \emph{arXiv preprint arXiv:2210.15865}.

\bibitem[{Liu et~al.(2021)Liu, Fan, Chen, Xu, and Yang}]{liu2021fate}
Yang Liu, Tao Fan, Tianjian Chen, Qian Xu, and Qiang Yang. 2021.
\newblock Fate: An industrial grade platform for collaborative learning with data protection.
\newblock \emph{J. Mach. Learn. Res.}, 22(226):1--6.

\bibitem[{Mangrulkar et~al.(2022)Mangrulkar, Gugger, Debut, Belkada, Paul, and Bossan}]{peft}
Sourab Mangrulkar, Sylvain Gugger, Lysandre Debut, Younes Belkada, Sayak Paul, and Benjamin Bossan. 2022.
\newblock Peft: State-of-the-art parameter-efficient fine-tuning methods.
\newblock \url{https://github.com/huggingface/peft}.

\bibitem[{McMahan et~al.(2017)McMahan, Moore, Ramage, Hampson, and y~Arcas}]{mcmahan2017communication}
Brendan McMahan, Eider Moore, Daniel Ramage, Seth Hampson, and Blaise~Aguera y~Arcas. 2017.
\newblock Communication-efficient learning of deep networks from decentralized data.
\newblock In \emph{Artificial intelligence and statistics}, pages 1273--1282. PMLR.

\bibitem[{OpenAI(2022)}]{Chatgpt}
OpenAI. 2022.
\newblock Chatgpt.

\bibitem[{Radford et~al.(2019)Radford, Wu, Child, Luan, Amodei, Sutskever et~al.}]{radford2019language}
Alec Radford, Jeffrey Wu, Rewon Child, David Luan, Dario Amodei, Ilya Sutskever, et~al. 2019.
\newblock Language models are unsupervised multitask learners.
\newblock \emph{OpenAI blog}, 1(8):9.

\bibitem[{Scao et~al.(2022)Scao, Fan, Akiki, Pavlick, Ili{\'c}, Hesslow, Castagn{\'e}, Luccioni, Yvon, Gall{\'e} et~al.}]{scao2022bloom}
Teven~Le Scao, Angela Fan, Christopher Akiki, Ellie Pavlick, Suzana Ili{\'c}, Daniel Hesslow, Roman Castagn{\'e}, Alexandra~Sasha Luccioni, Fran{\c{c}}ois Yvon, Matthias Gall{\'e}, et~al. 2022.
\newblock Bloom: A 176b-parameter open-access multilingual language model.
\newblock \emph{arXiv preprint arXiv:2211.05100}.

\bibitem[{Talmor et~al.(2018)Talmor, Herzig, Lourie, and Berant}]{talmor2018commonsenseqa}
Alon Talmor, Jonathan Herzig, Nicholas Lourie, and Jonathan Berant. 2018.
\newblock Commonsenseqa: A question answering challenge targeting commonsense knowledge.
\newblock \emph{arXiv preprint arXiv:1811.00937}.

\bibitem[{Team et~al.(2024)Team, Riviere, Pathak, Sessa, Hardin, Bhupatiraju, Hussenot, Mesnard, Shahriari, Ram{\'e} et~al.}]{team2024gemma}
Gemma Team, Morgane Riviere, Shreya Pathak, Pier~Giuseppe Sessa, Cassidy Hardin, Surya Bhupatiraju, L{\'e}onard Hussenot, Thomas Mesnard, Bobak Shahriari, Alexandre Ram{\'e}, et~al. 2024.
\newblock Gemma 2: Improving open language models at a practical size.
\newblock \emph{arXiv preprint arXiv:2408.00118}.

\bibitem[{Touvron et~al.(2023)Touvron, Lavril, Izacard, Martinet, Lachaux, Lacroix, Rozi{\`e}re, Goyal, Hambro, Azhar et~al.}]{touvron2023llama}
Hugo Touvron, Thibaut Lavril, Gautier Izacard, Xavier Martinet, Marie-Anne Lachaux, Timoth{\'e}e Lacroix, Baptiste Rozi{\`e}re, Naman Goyal, Eric Hambro, Faisal Azhar, et~al. 2023.
\newblock Llama: Open and efficient foundation language models.
\newblock \emph{arXiv preprint arXiv:2302.13971}.

\bibitem[{Wan et~al.(2024)Wan, Huang, Cai, Quan, Bi, and Shi}]{wan2024knowledge}
Fanqi Wan, Xinting Huang, Deng Cai, Xiaojun Quan, Wei Bi, and Shuming Shi. 2024.
\newblock Knowledge fusion of large language models.
\newblock \emph{arXiv preprint arXiv:2401.10491}.

\bibitem[{Wang et~al.(2019)Wang, Pruksachatkun, Nangia, Singh, Michael, Hill, Levy, and Bowman}]{wang2019superglue}
Alex Wang, Yada Pruksachatkun, Nikita Nangia, Amanpreet Singh, Julian Michael, Felix Hill, Omer Levy, and Samuel~R. Bowman. 2019.
\newblock Super{GLUE}: A stickier benchmark for general-purpose language understanding systems.
\newblock \emph{arXiv preprint 1905.00537}.

\bibitem[{Wang et~al.(2022)Wang, Mishra, Alipoormolabashi, Kordi, Mirzaei, Arunkumar, Ashok, Dhanasekaran, Naik, Stap et~al.}]{wang2022benchmarking}
Yizhong Wang, Swaroop Mishra, Pegah Alipoormolabashi, Yeganeh Kordi, Amirreza Mirzaei, Anjana Arunkumar, Arjun Ashok, Arut~Selvan Dhanasekaran, Atharva Naik, David Stap, et~al. 2022.
\newblock Benchmarking generalization via in-context instructions on 1,600+ language tasks.
\newblock \emph{arXiv preprint arXiv:2204.07705}, 2.

\bibitem[{Xia et~al.(2023)Xia, Gao, Zeng, and Chen}]{xia2023sheared}
Mengzhou Xia, Tianyu Gao, Zhiyuan Zeng, and Danqi Chen. 2023.
\newblock Sheared llama: Accelerating language model pre-training via structured pruning.
\newblock \emph{arXiv preprint arXiv:2310.06694}.

\bibitem[{Yang et~al.(2019)Yang, Liu, Cheng, Kang, Chen, and Yu}]{yang2019federated}
Qiang Yang, Yang Liu, Yong Cheng, Yan Kang, Tianjian Chen, and Han Yu. 2019.
\newblock Federated learning.
\newblock \emph{Synthesis Lectures on Artificial Intelligence and Machine Learning}, 13(3):1--207.

\bibitem[{Yang et~al.(2021)Yang, Tian, and Zhang}]{yang2021regularized}
Ruihong Yang, Junchao Tian, and Yu~Zhang. 2021.
\newblock Regularized mutual learning for personalized federated learning.
\newblock In \emph{Asian Conference on Machine Learning}, pages 1521--1536. PMLR.

\bibitem[{Yi et~al.(2023)Yi, Yu, Wang, and Liu}]{yi2023fedlora}
Liping Yi, Han Yu, Gang Wang, and Xiaoguang Liu. 2023.
\newblock Fedlora: Model-heterogeneous personalized federated learning with lora tuning.
\newblock \emph{arXiv preprint arXiv:2310.13283}.

\bibitem[{Zhang et~al.(2022{\natexlab{a}})Zhang, Roller, Goyal, Artetxe, Chen, Chen, Dewan, Diab, Li, Lin et~al.}]{zhang2022opt}
Susan Zhang, Stephen Roller, Naman Goyal, Mikel Artetxe, Moya Chen, Shuohui Chen, Christopher Dewan, Mona Diab, Xian Li, Xi~Victoria Lin, et~al. 2022{\natexlab{a}}.
\newblock Opt: Open pre-trained transformer language models.
\newblock \emph{arXiv preprint arXiv:2205.01068}.

\bibitem[{Zhang et~al.(2018)Zhang, Xiang, Hospedales, and Lu}]{zhang2018deep}
Ying Zhang, Tao Xiang, Timothy~M Hospedales, and Huchuan Lu. 2018.
\newblock Deep mutual learning.
\newblock In \emph{Proceedings of the IEEE conference on computer vision and pattern recognition}, pages 4320--4328.

\bibitem[{Zhang et~al.(2022{\natexlab{b}})Zhang, Yang, Dai, Qu, and Xu}]{zhang2022federated}
Zhuo Zhang, Yuanhang Yang, Yong Dai, Lizhen Qu, and Zenglin Xu. 2022{\natexlab{b}}.
\newblock When federated learning meets pre-trained language models' parameter-efficient tuning methods.
\newblock \emph{arXiv preprint arXiv:2212.10025}.

\bibitem[{Zhao et~al.(2022)Zhao, Du, Li, Li, and Liu}]{zhao2022reduce}
Haodong Zhao, Wei Du, Fangqi Li, Peixuan Li, and Gongshen Liu. 2022.
\newblock Reduce communication costs and preserve privacy: Prompt tuning method in federated learning.
\newblock \emph{arXiv preprint arXiv:2208.12268}.

\bibitem[{Zhu et~al.(2012)Zhu, Wang, Zhong, Liu, Li, and Yang}]{zhu2012discovering}
Yin Zhu, Xiao Wang, Erheng Zhong, Nathan Liu, He~Li, and Qiang Yang. 2012.
\newblock Discovering spammers in social networks.
\newblock In \emph{proceedings of the AAAI conference on artificial intelligence}, volume~26, pages 171--177.

\end{thebibliography}

\appendix

\section{FedMKT Workflow}
\label{sec:appendix-workflow}
 The workflow of FedMKT is outlined as follows:

\begin{enumerate}

\item In the $t$-th communication round, the $K$ clients train their respective LoRA adapters using their private data. This step allows the clients to adapt their models to their specific data distributions.

\item After local training, each client $k$ computes a knowledge set $\mathcal{S}_k=\{l^i_{k}, p^i_{k}\}_{i=1}^N$ consisting of loss-logit pairs through its local model based on the public dataset.

\item Each client $k$ upload $\mathcal{S}_k$ to the server. 

\item  On the server side, token alignment is performed from the SLMs to the LLM, guaranteeing compatibility between the SLMs and the LLM.

\item  On the server side, knowledge is selected from the SLMs to the LLM according to Algorithm \ref{alg:dualmince}.

\item  On the server side, knowledge is transferred from the SLMs to the LLM based on the selected knowledge.

\item Once the knowledge transfer from SLMs to LLM is completed on the server, the server then computes a knowledge set $\mathcal{S}_0=\{l^i_{0}, p^i_{0}\}_{i=1}^N$ consisting of loss-logit pairs through LLM based on the public dataset.

\item The server disseminates $\mathcal{S}_0$ to all the clients.

\item  On the client side, the token alignment flow reverses, and token alignment is performed from the LLM to SLMs.

\item  On the client side, knowledge is selected from the LLM to each client SLM according to Algorithm \ref{alg:dualmince}.

\item  On the client side, knowledge is transferred from the LLM to each client SLM based on the selected knowledge.

\end{enumerate}

\section{Implementation Details of Token Alignment}
\label{sec:appendix-MinED}

\begin{figure}[htb]
  \centering
  \includegraphics[width=0.98\linewidth]{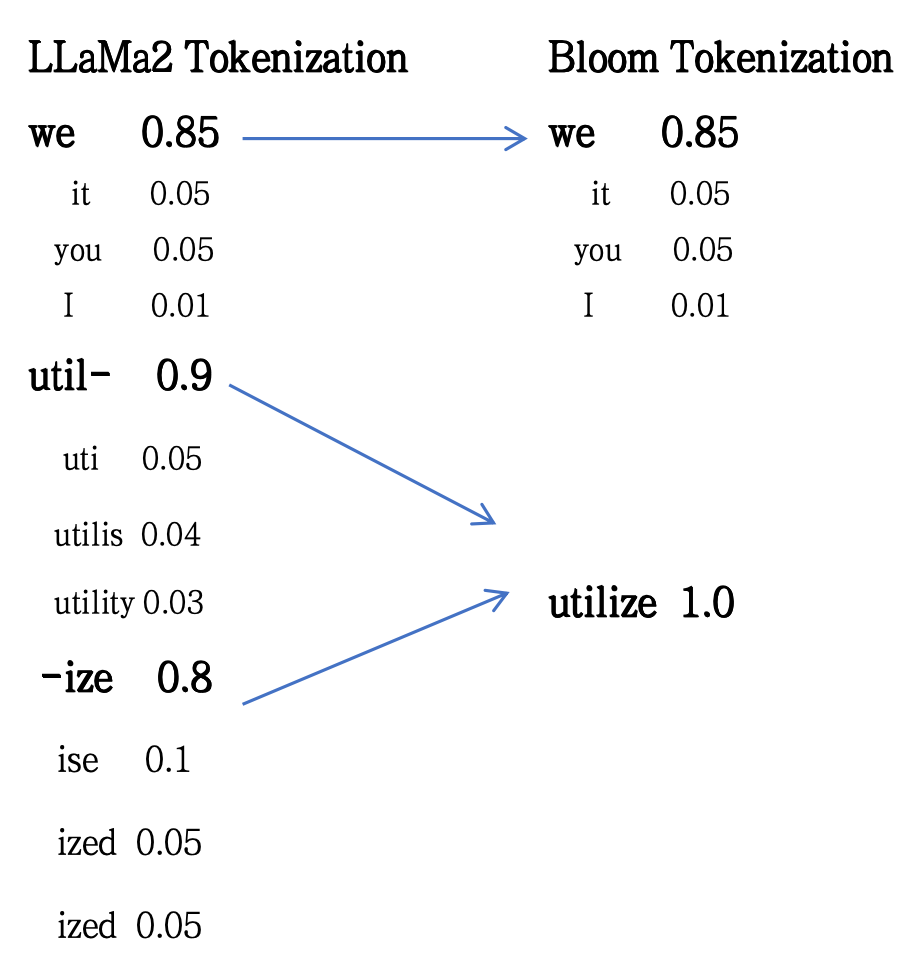}
  \caption{
 An example for token alignment via MinED.
  }
  \label{fig:tokenalign}
\end{figure}

In our work, we engage in a bidirectional token alignment procedure, encompassing the alignment of SLM tokens with their corresponding LLM tokens, and vice versa. Both alignments adhere to a similar methodology. Presently, we shall elaborate on the process of aligning LLM tokens with their matching SLM tokens.
To map the predicted token logits from the LLaMa2-7B (LLM) model to the Bloom-1.1B (SLM) model, several steps must be undertaken. The detailed process is as follows:

\begin{enumerate}
    \item Building an Optimal Vocabulary Mapping Table:
     
   \begin{enumerate}
   \item For each token in the LLaMa2 vocabulary, iterate through the Bloom vocabulary.

   \item Use \textit{minimum edit distance(MinED)} as a similarity measure to find the closest token in the Bloom vocabulary to the token in the LLaMa2 vocabulary.
   The recursion function for \textit{MinED} in dynamic programming is elaborated in \cite{fu2023specializing}.
   
   \item If there are multiple token with the same minimum edit distance, choose the one with the lexicographically smallest order.

   \item Save this mapping relationship in the optimal vocabulary mapping table.

   \end{enumerate}

 \item Tokenization and Alignment:

   \begin{enumerate}
  \item Tokenize the sentence "we utilize the dynamic programming approach to align tokens" using both the LLaMa2 and Bloom tokenizers.

  \item To align the two tokenization results and determine the optimal matching path, we utilize a dynamic programming algorithm. As an illustration, consider the tokenization outputs from LLaMa2 and Bloom. LLaMa2's tokenization yields: ['we', 'util', 'ize', 'the', 'dynamic', 'programming', 'approach', 'to', 'align', 'tokens']. In contrast, Bloom's tokenization produces: ['we', 'utilize', 'the', 'dynamic', 'programming', 'approach', 'to', 'align', 'tokens']. In this instance, seven terms from LLaMa2 align perfectly with those from Bloom, such as "we" and "dynamic". Notably, the LLaMa2 tokens 'util' and 'ize' collectively map to the single Bloom token 'utilize'. In scenarios where multiple tokens align to one, like the 2-to-1 case of 'util' and 'ize' mapping to 'utilize', we consider 'utilize' as a match for 'util' based on an optimal vocabulary mapping. Figure \ref{fig:tokenalign} illustrates an example of token alignment between LLaMa2 and Bloom. 

    \end{enumerate}

 \item Logits Mapping:
    
    \begin{enumerate}
   \item Iterate through each token $t_t$ in the Bloom tokenization result.

  \item For each $t_t$, check if it uniquely matches a token $t_s$ in the LLaMa2 tokenization result.

   \item If $t_t$ uniquely matches $t_s$, then for each token $t_p$ in the Top-$K$ predicted token of $t_s$ from LLaMa2 and its corresponding logit $logit_p$: Find the position $pos$ in the Bloom vocabulary that corresponds to $t_p$ using the optimal vocabulary mapping table. If $pos$ has not been assigned a value before, copy $logit_p$ to the corresponding position in the Bloom logits distribution matrix $logit_t$.

   \item If $t_t$ does not have a unique match, generate one-hot logits for $t_t$.
   \end{enumerate}
   
   \item  Processing the Results:
\begin{enumerate}
  \item Ultimately, each token $t_t$ in Bloom will have a corresponding logits distribution matrix $logit_t$.

   \item These logits can be directly used for subsequent training in the Bloom model.

   \end{enumerate}
\end{enumerate}

\section{Computation and Communication Complexity}
\label{sec:appendix-Computation}
One of the key advantages of FedMKT is its computational efficiency. By leveraging PEFT, the framework significantly reduces the number of parameters that need to be updated during fine-tuning. For instance, it consumes just 0.12\% of the computational cost associated with fine-tuning all parameters in OPT-1.3B when using FedMKT.
This leads to faster training times and reduced computational requirements, making it more feasible to fine-tune LLM and SLMs in a federated learning setting. 

In terms of communication complexity, FedMKT minimizes the amount of data exchanged between clients and the server. Instead of transmitting entire models(For example, OPT-1.3B is about 1.3B floating-point numbers), clients only share the output logits and corresponding cross-entropy losses of the public dataset with the server. 
Suppose there are $N=1000$ public text samples with a text sequence length of $S = 512$ and a top token size of $K = 16$. The communication cost, denoted as $Cost_{com}$, would be calculated as follows: $Cost_{com} = N * S * K = 1000 * 512 * 16 = 8M$ floating-point numbers. This approach reduces communication overhead, allowing for more efficient data transmission and enhancing scalability in federated learning scenarios. 

\section{More on Experimental Details}
\label{sec:appendix-Experimental}
\subsection{Hyperparameter Settings}

\textbf{LoRA Parameters.} We utilized the PEFT\cite{peft} library with the following configurations: r=8, lora\_alpha=16, lora\_dropout=0.05. 

\textbf{Common Parameters for LLM and SLMs.} We set batch\_size=4, used the AdamW optimizer with adam\_beta1=0.9 and adam\_beta2=0.95. The warmup\_ratio was set to 0.008, the weight\_decay was 0.1, max\_grad\_norm was 1.0. The $\lambda$ was 0.9. The number of training rounds for all data is within 10 and the number of training rounds for different datasets may be different.

\textbf{LLM Parameters.} During distillation, the local epoch R was set to 1. The learning rates $\eta_\omega$ were specified as 3e-5 for the datasets RTE/WIC/BoolQ/CQA/ARC-C/DialogSum/S-NI, and 2e-5 for ARC-E.

\textbf{SLM Parameters.} During training for the four clients, the local epoch E was set to 1. The learning rates $\eta_\theta$ were as follows: for "OPT-1.3B", $\eta_\theta$=3e-5; for "GPT-2-xlarge", $\eta_\theta$=3e-4; for "Bloom-1.1B", $\eta_\theta$=3e-5; and for "LLaMa-2-1.3B", the same learning rates as for the LLM were used.

\subsection{Data Splitting}

For the datasets RTE/WIC/BoolQ/CQA/ARC-E/ARC-C/DialogSum, we randomly split the training data into five equal parts, with one part serving as the public dataset and the remaining four parts as private dataset for the four clients. All these datasets(including train, validate, test) were downloaded from HuggingFace\cite{lhoest-etal-2021-datasets}. For the S-NI dataset, we first processed the data using minillm\cite{gu2023minillm} to retain samples with an output length greater than or equal to 11. From this processed data, we randomly selected 300 samples as the evaluation dataset. The remaining data was then split into five equal parts, with one part serving as the public dataset and the other four parts as private data for the four clients.

\subsection{Machine Configuration}

The experiments were conducted on machines equipped with either 4 Nvidia V100 32G or 8 Nvidia V100 32G GPUs.

\end{document}